\documentclass{article}

\usepackage{microtype}
\usepackage{graphicx}
\usepackage{subfigure}
\usepackage{booktabs} 
\usepackage{amsmath}
\usepackage{dsfont}

\usepackage{hyperref}
\usepackage{bm}
\usepackage{bbm}
\usepackage{enumitem}

\usepackage{xspace}
\usepackage{amssymb}

\usepackage[accepted]{icml2020}

\newcommand{\dmv}{\textsc{DMV-Full}\xspace}
\newcommand{\census}{\textsc{Census}\xspace}
\newcommand{\kdd}{\textsc{KDD}\xspace}
\newcommand{\dryad}{\textsc{Dryad-URLs}\xspace}

\DeclareMathOperator*{\PSample}{ProgressiveSampling}
\DeclareMathOperator*{\Dom}{Dom}
\DeclareMathOperator*{\Embed}{Embed}
\DeclareMathOperator*{\Encode}{Encode}

\newcommand{\mask}{\text{MASK}}

\icmltitlerunning{Variable Skipping for Autoregressive Range Density Estimation}

\begin{document}

\twocolumn[
\icmltitle{Variable Skipping for Autoregressive Range Density Estimation}



\icmlsetsymbol{equal}{*}

\begin{icmlauthorlist}
\icmlauthor{Eric Liang*}{cal}
\icmlauthor{Zongheng Yang*}{cal}
\icmlauthor{Ion Stoica}{cal}
\icmlauthor{Pieter Abbeel}{cal,cov}
\icmlauthor{Yan Duan}{cov}
\icmlauthor{Xi Chen}{cov}
\end{icmlauthorlist}

\icmlaffiliation{cal}{EECS, UC Berkeley, Berkeley, California, USA}
\icmlaffiliation{cov}{covariant.ai, Berkeley, California, USA}

\icmlcorrespondingauthor{Eric Liang}{ericliang@berkeley.edu}
\icmlcorrespondingauthor{Zongheng Yang}{zongheng@berkeley.edu}

\icmlkeywords{Machine Learning, ICML}

\vskip 0.3in
]



\printAffiliationsAndNotice{\icmlEqualContribution} 

\begin{abstract}
Deep autoregressive models compute \textit{point} likelihood estimates of individual data points.
However, many applications (i.e., database cardinality estimation) require estimating \textit{range} densities, a capability that is under-explored by current neural density estimation literature.
In these applications, fast and accurate range density estimates over high-dimensional data directly impact user-perceived performance.
In this paper, we explore a technique, \textit{variable skipping}, for accelerating range density estimation over deep autoregressive models.
This technique exploits the sparse structure of range density queries to avoid sampling unnecessary variables during approximate inference.
We show that variable skipping provides 10-100$\times$ efficiency improvements when targeting challenging high-quantile error metrics, enables complex applications such as text pattern matching, and can be realized via a simple data augmentation procedure without changing the usual maximum likelihood objective.
\end{abstract}

\section{Introduction}
Deep autoregressive (AR) models have achieved state-of-the-art density estimation results in image, video, and audio~\cite{pixelcnnpp, wavenet, pixelcnn,sparse_transformer,gpt2,weissenborn2019scaling}.
Recent work has applied them to domains traditionally outside of machine learning, such as physics~\cite{ar_physics}, protein modeling~\cite{tape}, and database query optimization~\cite{naru,neurocard}.
These use cases have surfaced the need for complex inference capabilities from deep AR models.
For example, the database cardinality estimation task reduces to estimating the density mass occupied by sets of variables under sparse \emph{range constraints}. 
In this problem, the database optimizer probes the fraction of records satisfying a query of high-dimensional constraints, e.g., $\text{Pr}(\textsf{age} > 35 \,\,\&\&\,\, \textsf{salary} < 50K)$, and relies on accurate estimates to pick performant query execution strategies.
\begin{figure}[t!]
\centering
      \includegraphics[trim=.6cm .6cm .5cm .5cm, clip,width=\columnwidth]{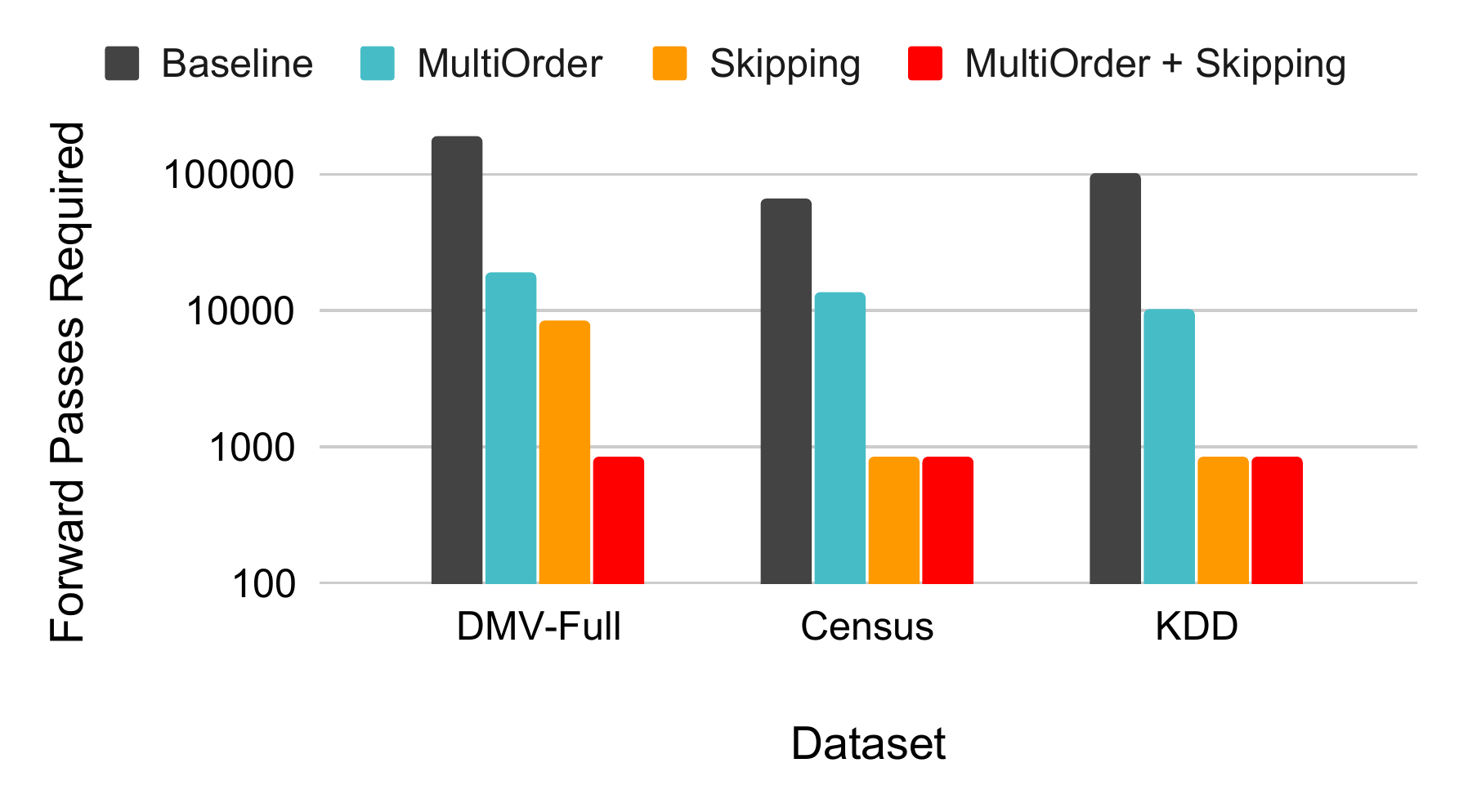}
\vspace{-.2in}
\caption{Approximate number of model forward passes required to achieve single-digit inference error at the \textit{99th quantile}. Y-axis shown in \textit{log scale}, lower is better. Variable skipping provides 10-100$\times$ compute
  savings
  for challenging high-quantile error targets. Refer to the Evaluation section for full results. \label{fig:summary}}
\end{figure}

In this paper, we call for attention to such \emph{range density estimation} problems 
in the context of deep AR models.
Given rapid advances in model capabilities, fast and accurate range density estimation has broad potential applicability to a number of domains, including databases, text processing, and inpainting (Section~\ref{sec:apps}).

Range density estimation involves two related challenges:
\setlist{nolistsep}
\begin{itemize}[noitemsep]
\item \textbf{Marginalization}: the handling of unconstrained variables, and 
\item \textbf{Range Constraints}: variables that are constrained to a specific \textit{range} or \textit{subset} of values. 
\end{itemize}
Exact inference or integration over the query region takes time exponential in the number of dimensions---a cost too high for all but the tiniest problems.
Further, both marginalization and range constraints are difficult to implement on top of AR models since they are only trained to provide point density estimates.
This motivates the use of approximate inference algorithms such as 
recently proposed by \cite{naru},
which show that AR models can significantly improve on the state-of-the-art in range estimation accuracy while remaining competitive in latency. 

\begin{figure}[!t]
      \includegraphics[width=\columnwidth]{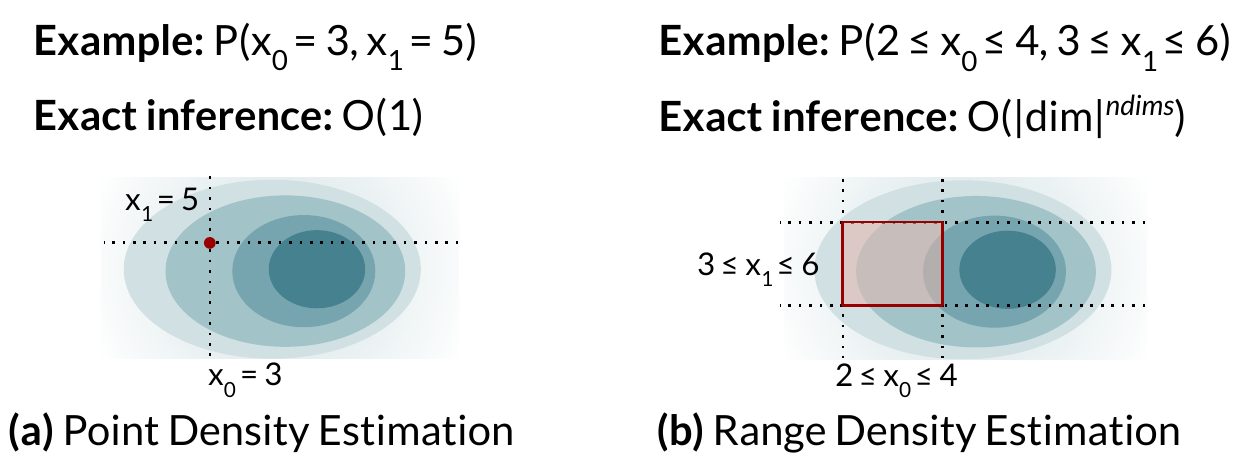}
\vspace{-.5cm}
    \caption{Comparison of point density and range density estimation. Naive marginalization to estimate range densities takes time proportional to the size of the query region (i.e., exponential in the number of dimensions of the joint distribution). \label{fig:point-range}}
\end{figure}

Building on prior work, we distill and evaluate a more general optimization for accelerating range density estimation termed \textit{variable skipping}.
The central idea is to exploit the sparsity of range queries, by avoiding sampling through the unconstrained dimensions (i.e., those to be marginalized over) during approximate inference.
A training-time data augmentation procedure randomly replaces some dimensions in the input with learnable marginalization tokens, which are trained to represent the \emph{absence} of those dimensions. During inference, the unconstrained dimensions take on these learned values instead of being sampled from their respective domains.

Variable skipping provides two key advantages.
First, by not needing to sample a concrete value for certain variables, the number of forward passes is significantly reduced from $O(|\text{Vars}|)$ (e.g., hundreds) to $O(|\text{ConstrainedVars}|)$ (e.g., a few).
Second, by avoiding sampling through the (potentially large) unconstrained region, it is possible to reduce the variance of the sampling-based estimator. We show that variable skipping realizes both advantages in practice (Figure \ref{fig:summary}).

Reducing the computation required for estimates can significantly impact the viability of model-based estimators for the aforementioned computer systems applications. For example, in database query optimization, cardinality estimation is typically run in the inner loop of a dynamic program~\cite{systemr}, and hence has to be executed many times in potentially unbatchable fashion. Further,
this process
must be re-run for each new query as it may have different variable constraints.
In this setting, reducing estimation costs from tens or hundreds of forward passes (i.e., the number of columns in a typical production database) to just a handful (i.e, the number of constraints in a typical range query) is critical for adoption.
Models that include rarely queried text columns (e.g., byte pair encoded, which exacerbates the problem) may benefit further still.


We start by discussing related work, then reviewing the previously proposed approximate inference algorithm~\cite{naru}, termed Progressive Sampling (Section \ref{sec:prog-sample}), which allows any trained autoregressive model
to efficiently compute range densities.
We then discuss an optimization, \emph{variable skipping}, which allows dimensions irrelevant to a query to be skipped over at inference time, greatly reducing or eliminating sampling costs (Section \ref{sec:variable-skipping}).
We show that, beyond accelerating range density estimation, variable skipping can enable related applications such as pattern matching and text completion.
Finally, we study the performance of variable skipping (Section \ref{sec:evaluation}).

The contributions of this paper are as follows:
\begin{enumerate}
\item We distill the more general concept of \textit{variable skipping}, a training and run-time optimization that greatly reduces the variance of range density estimates. 
\item To show its generality, we apply variable skipping to text models, which can then support applications such as pattern matching.
\item We evaluate the effectiveness of variable skipping across a variety of datasets, architecture, and hyperparameter choices, and compare with related techniques such as multi-order training.
\item To invite research on this under-explored problem, we open source our code and a set of range density estimation benchmarks on high-dimensional discrete datasets at \small{\url{https://var-skip.github.io}}.
\end{enumerate}

\subsection{Applications of Range Density Estimation}
\label{sec:apps}
Range density estimation is important for the following applications, among others:


\textbf{Database Systems:} 
A core primitive in database query optimizer is cardinality estimation~\cite{systemr}: given a query with user-defined predicates for a subset of columns, estimate the \emph{fraction} of records that satisfy the predicates. Applying AR models to cardinality estimation was the topic of \cite{naru}.

\textbf{Pattern Matching:} A regular expression can be interpreted as a dynamically unrolled predicate (i.e., a nondeterministic finite automata) \cite{hopcroft} over a series of character variables. Hence, its \emph{match probability} can be estimated in  the same way as a range query. Section \ref{subsec:text} shows how this can be realized with variable skipping.

\textbf{Completion and Inpainting:} While an AR model can be straightforwardly used to extend a prefix in the variable ordering, completing a missing value from the \textit{middle} of a sequence of variables requires sampling from the marginal distribution over missing values. We show that variable skipping  allows this to be done efficiently (Section \ref{subsec:completion}).

\section{Related Work}

\textbf{Density Estimation with Deep Autoregressive Models} have enjoyed vast interest due to their outstanding capability of modeling high-dimensional data (text, images, tabular).  Efficient architectures such as MADE~\cite{made} and ResMADE~\cite{resmade} have been proposed, and self-attention models (e.g., Transformer~\cite{transformer}) have underpinned recent new advances in language tasks.
Our work optimizes the approximate inference (of range density estimates) on top of such AR architectures.


\textbf{Masked Language Models}.  Our variable skipping learns special MASK tokens (Section~\ref{sec:variable-skipping}) by randomly masking inputs, which is similar to masked language models  such as BERT~\cite{bert} and CMLMs~\cite{cmlm}.  These models differ from AR models in optimization goals: they typically predict \emph{only} the masked tokens conditioned on present tokens, and may assume independence among the masked tokens.
We study deep AR models for two reasons: (1) our problem settings are in density estimation, and deep AR models have generally shown superior density modeling quality than other generative models; (2) the approximate inference procedure we study (Section~\ref{sec:prog-sample}) assumes access to autoregressive factors.


\noindent \textbf{Multi-Order Training} handles marginalization by training over many orders and invoking a suitable order (or an ensemble over available orders) during inference. This technique has appeared in NADE~\cite{uria13_deep_tract_densit_estim}, MADE~\cite{made}, XLNet~\cite{xlnet}, among others.
Variable skipping shares the same goal of efficiently handling marginalization.
These prior works have reported increased optimization difficulty as the number of orders to learn increases (some sample a fixed set of orders, while others keep sampling new orders).
In the latter case, we posit that the difficulty is due to adding $n!$ input variations; in contrast, variable skipping only extends the vocabulary of each dimension by a $\text{MASK}$ symbol, a relatively smaller increase in task difficulty. In Section~\ref{sec:evaluation}, we compare variable skipping against multi-order training, and show that they can be combined to further reduce errors.
\section{Range Density Estimation on Deep Autoregressive Models}
We model a finite set of $n$-dimensional data points $D = \{ \bm{x}^{(0)}, \bm{x}^{(1)}, \dots, \bm{x}^{(M)}\}$ as a discrete distribution using an autoregressive model $p_\theta(\bm{x})$, parameterized by $\theta$. The model is trained on $D$ using the maximum likelihood objective:
\begin{equation}
  \min_\theta\, \mathcal{L}(\theta) = -\mathop{\mathbb{E}}_{\bm{x} \sim D} \log p_{\theta}(\bm{x})
\end{equation}
where $p_{\theta}(\bm{x}) = \prod_{i} p_{\theta}({x}_i | \bm{x}_{< i})$ for each data point $\bm{x}$.

\textbf{Range density.}  We consider range queries of the form
\begin{equation}
  p_\theta(X_1 \in R_1, \dots, X_n \in R_n), \label{eq:range-query}
\end{equation}
where each region $R_i$ is a subset of the domain $\Dom(X_i)$.  This formulation encapsulates unconstrained dimensions, where we simply take $R_i = \Dom(X_i)$ (the whole domain).

\subsection{Background: Progressive Sampling}
\label{sec:prog-sample}
Exact inference of Equation~\ref{eq:range-query} is computationally efficient only for low dimensions or small domain sizes.  Approximate inference is required to scale its computation.

To solve this problem, \cite{naru} adapts classical forward sampling~\cite{pgm_textbook} for range likelihoods, yielding an unbiased approximate inference algorithm.
The algorithm works by drawing in-range samples and re-weighting each intermediate range likelihood.
Each in-range sample is drawn from the first dimension to the last (in the AR ordering).
As an example, consider estimating $p_\theta(X_1 \in R_1, X_2 \in R_2, X_3 \in R_3)$.
Progressive sampling draws $x_1^{MC} \sim p_\theta(X_1 | X_1 \in R_1)$ and stores $p_\theta(X_1 \in R_1)$---both tractable operations since a forward pass on the trained AR produces this single-dimensional distribution.  It performs another forward pass to obtain $p_\theta(X_2 | x_1^{MC})$, which then produces a sample $x_2^{MC}$ and the range likelihood $p_{\theta}(X_2 \in R_2 | x_1^{MC})$.  Lastly it obtains $p_{\theta}(X_3 \in R_3 | x_1^{MC}, x_2^{MC})$. It can be shown that the product of all $n$ range likelihoods, e.g.,
\[
p_\theta(X_1 \in R_1) 
p_{\theta}(X_2 \in R_2 | x_1^{MC}) 
p_{\theta}(X_3 \in R_3 | x_1^{MC}, x_2^{MC})
\]
is a valid Monte-Carlo estimate of the desired range density.

In the remainder of the paper, we invoke $\PSample(R_1, \dots, R_n)$ as a black-box estimator, although our variable technique (described next) can work with other estimators for Equation~\ref{eq:range-query}.

\section{Variable Skipping}
\label{sec:variable-skipping}

Variable skipping works by (1) training special marginalization tokens, $\mask_i$, for each dimension $i$; (2) at approximate inference, rewriting each unconstrained variable, e.g., $X_i \in \Dom(X_i)$, into a constrained variable with the singleton range, $X_i \in \{ \mask_i \}$. The training process can be interpreted as dropout of the input, or as data augmentation. 

\begin{figure}[t]
\centering

      \includegraphics[width=\columnwidth]{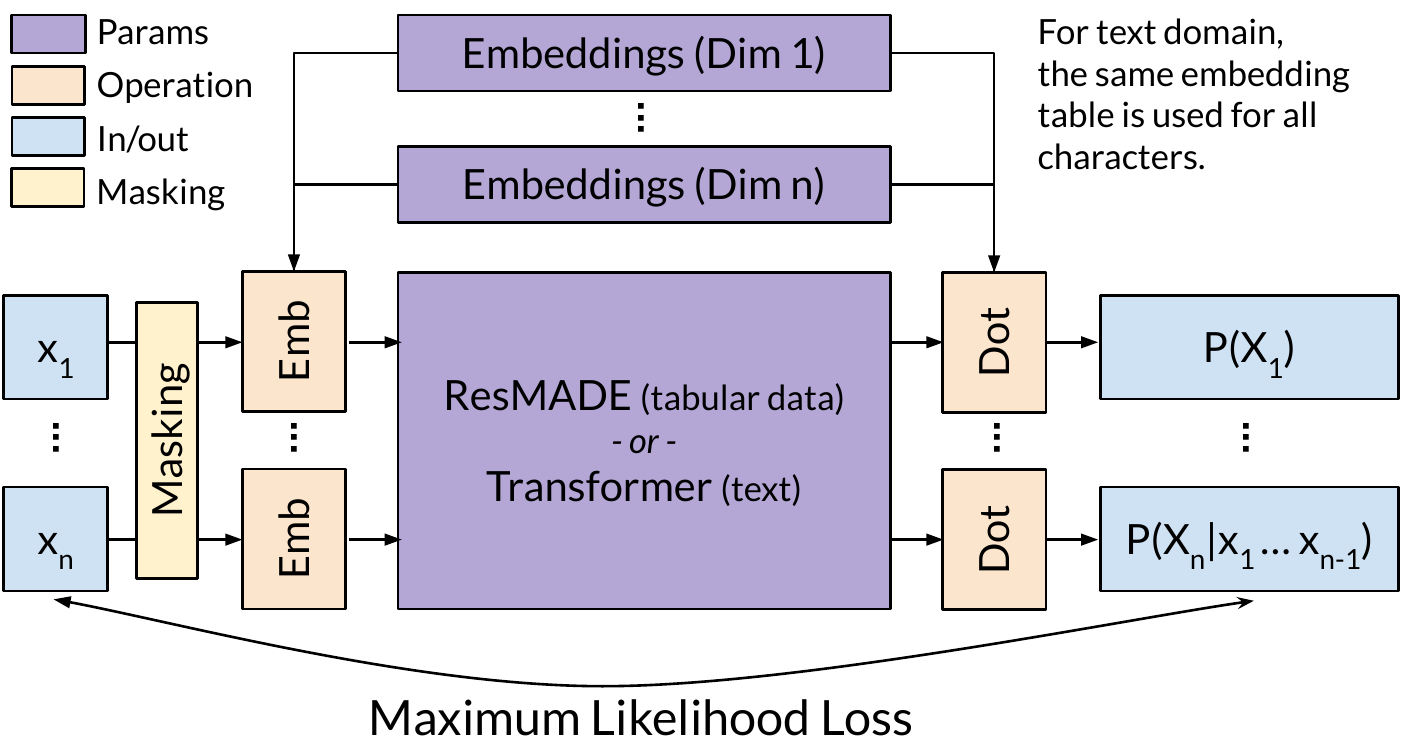}
\vspace{-.2in}
    \caption{Model architecture. \label{fig:architecture}}
\end{figure}

\textbf{Architecture}. We assume a model architecture shown in Figure~\ref{fig:architecture}: the input layer, an autoregressive core, and the output layer.
For the autoregressive core,  we use ResMADE~\cite{resmade} for tabular data and an autoregressive Transformer~\cite{transformer} (encoder only with correct masking) for text data.  At the input layer, we embed each data point using a per-dimension trainable embedding table, denoted by $\Embed()$.  (For text, we tie the embeddings across all dimensions since they share the same character vocabulary.)  The output layer dots the hidden features with the input embeddings to produce logits.

\textbf{Training-time input masking.}   First, we add a special token $\mask_i$ to each dimension $i$'s vocabulary.  
For each input $\bm{x}$ we uniformly draw the number of masked dimensions $n_{\text{mask}} \sim [0, |\bm{x}|)$, then sample the $n_{\text{mask}}$ positions to mask, $X_{\text{mask}}$.  For position $i \in X_{\text{mask}}$, we replace the original representation, $\Embed(x_i)$, by the masked representation, $\Embed(\mask_i)$:
\begin{equation}
\label{eq:mask}
\begin{split}
  \Encode(x_i) = & \Embed(x_i) \mathds{1}(i \notin X_{\text{mask}}) \\
  & + \Embed(\mask_i) \mathds{1}(i \in X_{\text{mask}}).
\end{split}
\end{equation}
Importantly, the objective remains the MLE for \emph{all} autoregressive factors: we train the parameters to predict the \emph{original values} at each dimension, given \emph{a mix of original and masked information} at previous dimensions.  In other words, we minimize the negative log-likelihood
\begin{equation}
\begin{split}
   -\log p_\theta(x_i | \bm{x}_{<i})   =  -\log p_\theta\left (x_i | \forall j < i, \Encode(x_j) \right )
\end{split}
\end{equation}
over all dimension $i$. Conditioning on the mask tokens ensures that those representations are trained. Since we do not alter the \textit{output targets} and the mask positions are chosen independently of the data, no bias is introduced.

\textbf{Infer-time skipping.}  Given a range query (Equation~\ref{eq:range-query}),
we look for each unconstrained dimension and replace its domain with a singleton set of its marginalization token:
\begin{equation}
\begin{split}
&p_\theta(\dots, X_i \in R_i = \Dom(X_i), \dots) \\
 \rightarrow \quad & p_\theta(\dots, X_i \in R_i^{'} = \{ \text{MASK}_i \}, \dots)
\end{split}
\end{equation}
We then invoke $\PSample()$ which would thus \emph{skip} the sampling for those dimensions.

\textbf{Example.} Suppose we have an AR model trained over the autoregressive ordering $\{\text{age}, \text{salary}, \text{city}\}$, and want to draw a sample from $p_\theta(\text{city} | \text{salary} > 50\text{k})$.

\begin{itemize}
\item Without skipping, first we draw $x_1\sim p_{\theta}(\text{age})$, then $x_2\sim p_{\theta}(\text{salary}|\text{age}=x_1,\text{salary} > 50\text{k})$, and finally $x_3 \sim p_{\theta}(\text{city}|\text{age}=x_1, \text{salary}=x_2)$.

\item With skipping, we can directly sample $x_2\sim p_{\theta}(\text{salary}|\text{age}=\text{MASK}_1,\text{salary} > 50\text{k})$, followed by $x_3 \sim p_{\theta}(\text{city}|\text{age}=\text{MASK}_1, \text{salary}=x_2)$.
\end{itemize}



\begin{figure}[t]
\centering
      \includegraphics[width=\columnwidth]{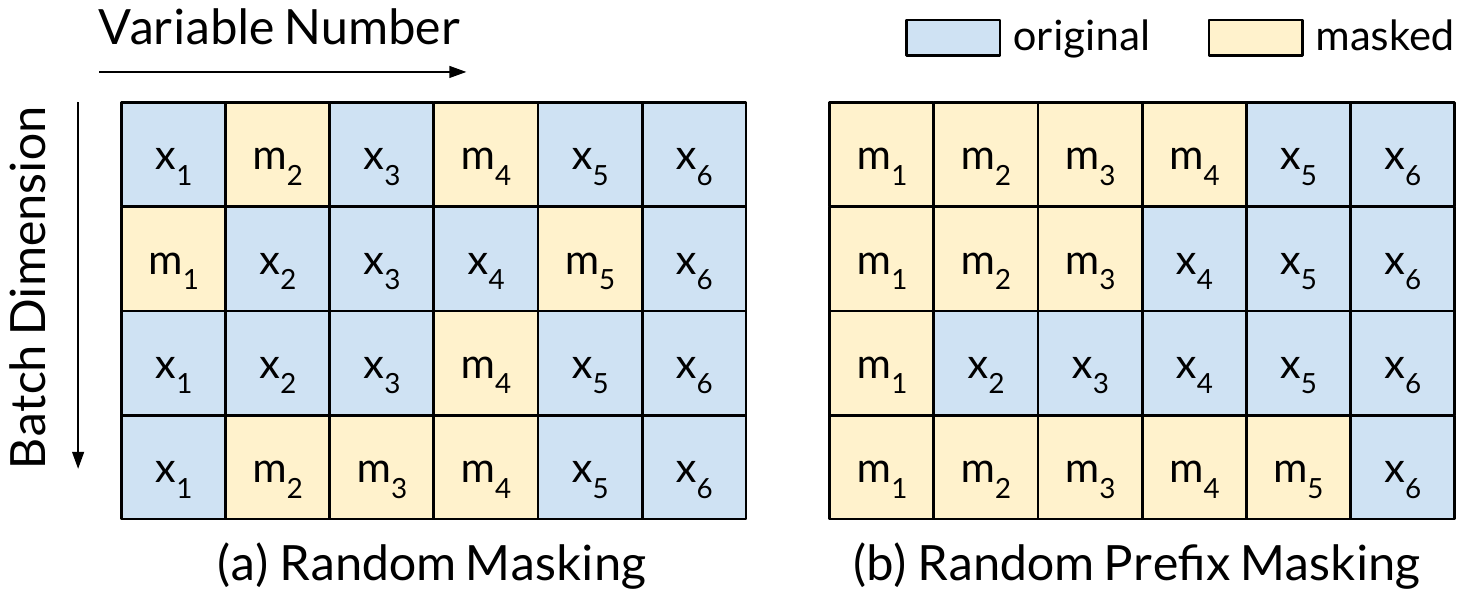}
\vspace{-.2in}
\caption{Masking strategies (Section~\ref{sec:variable-skipping}). \textit{(a)} For tabular data, we randomly sample the dimensions to mask out for each row. \textit{(b)} For text, we mask a random prefix of each string, exploiting the natural left-to-right ordering.
    \label{fig:masking}
    }
\end{figure}




\subsection{Prefix Skipping for Text Pattern Matching}
\label{subsec:text}

Any regex can be implemented as a nondeterministic finite automata (NFA) \cite{hopcroft}, which takes a stream of characters and determines acceptable next characters. We can use progressive sampling with any regex, treating its NFA like a dynamically unrolled predicate. For example, consider the regex $\mathtt{[at](c+|i+)e}$. Possible matches include $\mathtt{ace}$, $\mathtt{aiie}$, and $\mathtt{tiiie}$. Progressive sampling would work as follows: first we sample $x_1 \in \{a, t\}$, then $x_2 \in \{c, i\}$. Depending on whether $x_2=c$ or $x_2=i$, third we either sample $x_3 \in \{c, e\}$ or $\{i, e\}$ (this is the ``dynamic'' part), and so on. By retaining an NFA per sample, we obtain an estimate of the overall match probability.

However, this naive formulation is inefficient when there are long unconstrained sequences. Consider the regex $\mathtt{.*icml.*}$, intended to match any string containing the token $\mathtt{icml}$.
The probability of sampling a random prefix from an AR model matching this is vanishingly small---perhaps millions of samples for a hit.
To avoid this, we can try to \textit{skip over} sequences of unconstrained characters and compute the probability of $\mathtt{icml}$ at specific offsets directly.
All that would remain is sampling forward through the remainder of the variables to avoid double counting duplicate occurrences of the token.
Using $m(x_i)$ to denote a string match at position $i$, and $m(x_{>i})$ the existence of a match at any position $>i$, the match probability is approximated as:
\begin{equation}
\begin{split}
   P_{\text{match}} = & \sum_{i=1}^{n} P(m(x_{i})) \cdot (1 - P(m(x_{>i})|m(x_{i})) \\
   \approx & \sum_{i=1}^{n} p_{\theta}(m(x_i)| \{ x_j = \text{MASK}_j : j < i \}) \\
  &  (1 - p_{\theta}(m(x_{>i})|\{ x_j = \text{MASK}_j : j < i \},m(x_i))
\end{split}
\end{equation}
Due to the need for masking contiguous prefixes, the model is trained with \textit{random prefix masking} (Figure \ref{fig:masking}) to allow such contiguous characters to be skipped.
We show the effectiveness of this strategy in Section \ref{subsec:texteval}, which implements simple pattern queries over an AR Transformer model.

\subsection{Other Mask Patterns}
\label{subsec:completion}

Finally, we note that more structured mask patterns can be used, such as sub-sequences in text or random patches in images \cite{dupont2018probabilistic}. This allows for marginalization over complex subsets of dimensions with potential applicability to not only sample variables given prefixes of the AR ordering (i.e., from $P(x_i|x_1,\ldots,x_{i-1})$) but also variables \textit{later on}, i.e., from $P(x_i|x_1,\ldots,x_{i-1},x_{i+k},\ldots,x_N)$ by marginalizing over $\{x_{i+1},\ldots,x_{i+k-1}\}$. We leave investigation of these potential applications to future work.

\section{Evaluation}
\label{sec:evaluation}
\vspace{-.1cm}
Our evaluation investigates the following questions:

\begin{enumerate}
\item How much does variable skipping improve estimation accuracy compared to baselines, and how is this impacted by the sampling budget?


\item Can variable skipping be combined with multi-order training to further improve accuracy?

\item To what extent do hyperparameters such as the model capacity and mask token distribution impact the effectiveness of variable skipping?

\item Can variable skipping be applied to related domains such as text, or is it limited to tabular data?
\end{enumerate}

Overall, we find that variable skipping robustly improves estimation accuracy across a variety of scenarios. Given a certain target accuracy, skipping reduces the required compute cost by one to two orders of magnitude.

\begin{table}[tp]
  \small
  \centering
\caption{Datasets used in evaluation. ``Domain'' refers to the range of distinct values per table column (i.e., \dryad~contains 78 different character values). \label{table:datasets}}
\vspace{.1cm}
\begin{tabular}{@{} l l l l l l @{}} \toprule
Dataset & Rows & Cols & Domain & Type  \\ \midrule
\textsc{\dmv} & 11.6M & 19 & 2--32K & Discrete \\ 
\textsc{\census} & 2.5M & 67 & 2--18 & Discrete \\ 
\textsc{\kdd} & 95K & 100 & 2--896 & Discrete \\ 
\textsc{\dryad} & 2.4M & 100 & 78 & Text \\
\bottomrule
\end{tabular}
\end{table}

\subsection{Datasets}

We use the following public datasets in our evaluation, also summarized in Table \ref{table:datasets}. When necessary, we drop columns representing continuous data. We consider supporting continuous variables an orthogonal issue, and limit our evaluation to discrete domains:

\textbf{\dmv}~\cite{dmv-dataset}. Dataset consisting of vehicle
registration information in New York (i.e., attributes like vehicle class, make, model, and color). We use all columns except for the unique vehicle ID ({\sf VIN}).
This dataset was also used in \cite{naru}, but there it was restricted to 11 of the smaller columns.

\textbf{\kdd}~\cite{uci}. KDD Cup 1998 Data. We used the first hundred columns, sans {\sf noexch}, {\sf zip}, and {\sf pop901-3}, which were especially high-cardinality. This leaves 100 discrete integer domains with 2 to 896 distinct values each.

\textbf{\census}~\cite{uci}. The US Census Data (1990) Data Set, which consists of a 1\% sample made publicly available. We use all available columns, which range from 2 to 18 distinct values each.

\textbf{\dryad}~\cite{dryad}. For text domain experiments, we use this small dataset of 2.4M URLs, each truncated to 100 characters. This dataset was chosen to emulate a plausible \texttt{STRING} column in a relational database.

\begin{table}[tp]
  \small
  \centering
\caption{Hyperparameters for all experiments. We used a ResMADE for tabular data, and a Transformer for text. \label{table:hyperparameters}}
\vspace{.1cm}
\begin{tabular}{l l} \toprule
Hyperparameter & Value  \\ \midrule
Training Epochs & 20 (200 for KDD) \\
Batch Size & 2048 \\
Architecture & ResMADE \\
Residual Blocks & 3\\
Hidden Layers / Block & 2 \\
Hidden Layer Units & 256 \\
Embedding Size & 32 \\
Optimizer & Adam \\
Learning Rate & 5e-4 \\
Learning Rate Warmup & 1 epoch \\
Mask Probability & $\sim \text{Uniform}[0, 1)$ \\
\midrule
Transformer Num Blocks & 8 \\
Transformer MLP Dims ($d_{\text{ff}}$) & 256 \\
Transformer Embed Size ($d_{\text{model}}$) & 32 \\
Transformer Num Heads & 4 \\
Transformer Batch Size & 512 \\
\bottomrule
\end{tabular}
\end{table}

\begin{figure*}[ht]

\centering
\includegraphics[height=.9cm]{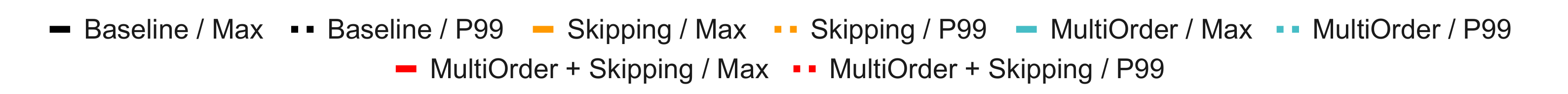}
\vspace{-.2cm}
\includegraphics[height=2.48cm]{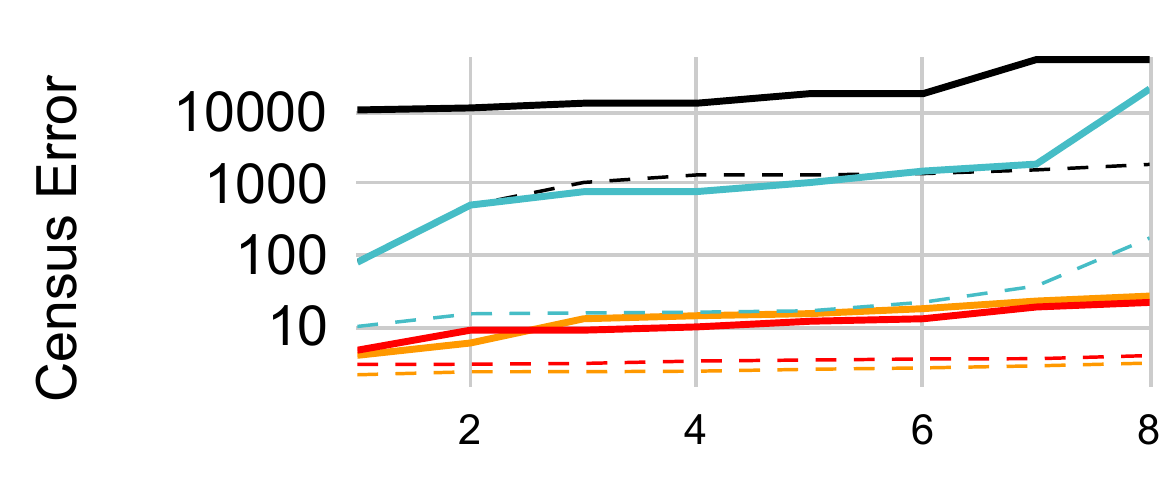}
\includegraphics[height=2.48cm]{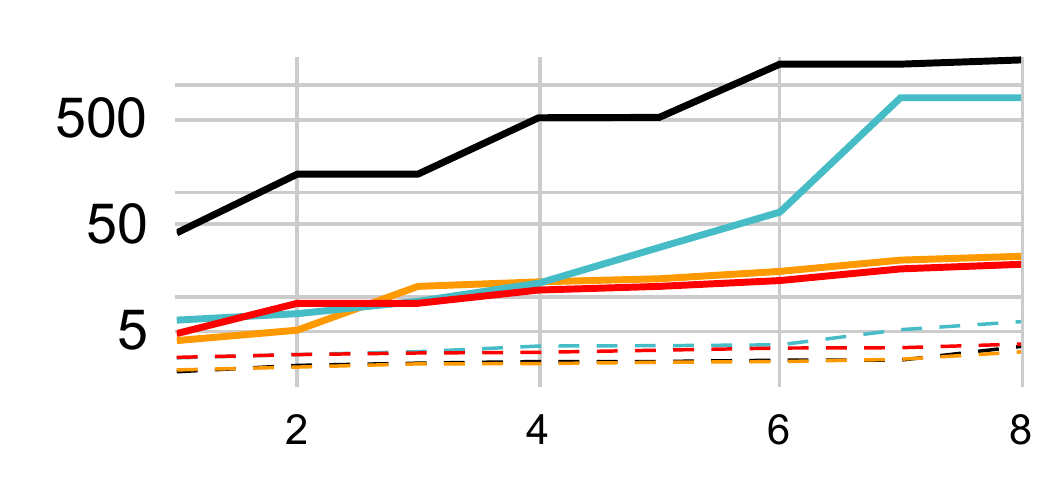}
\includegraphics[height=2.48cm]{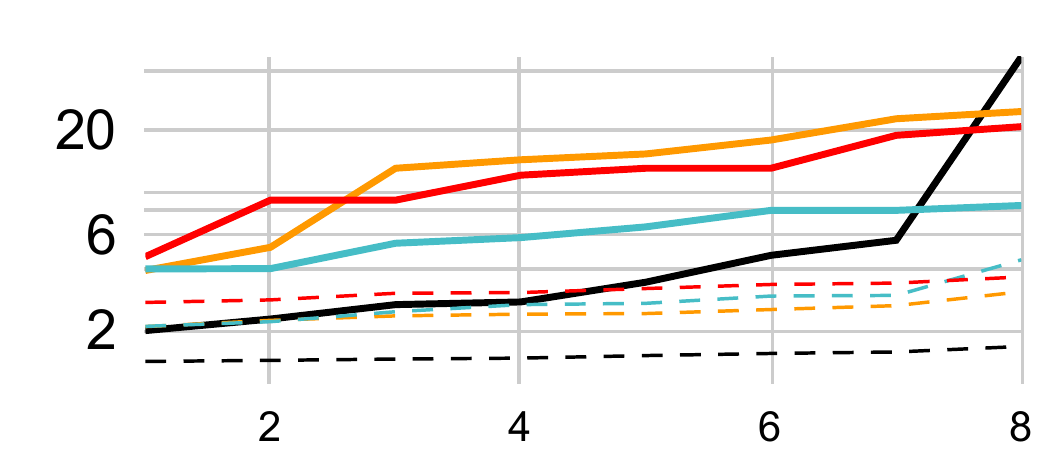}
\vspace{-.2cm}
\includegraphics[height=2.48cm]{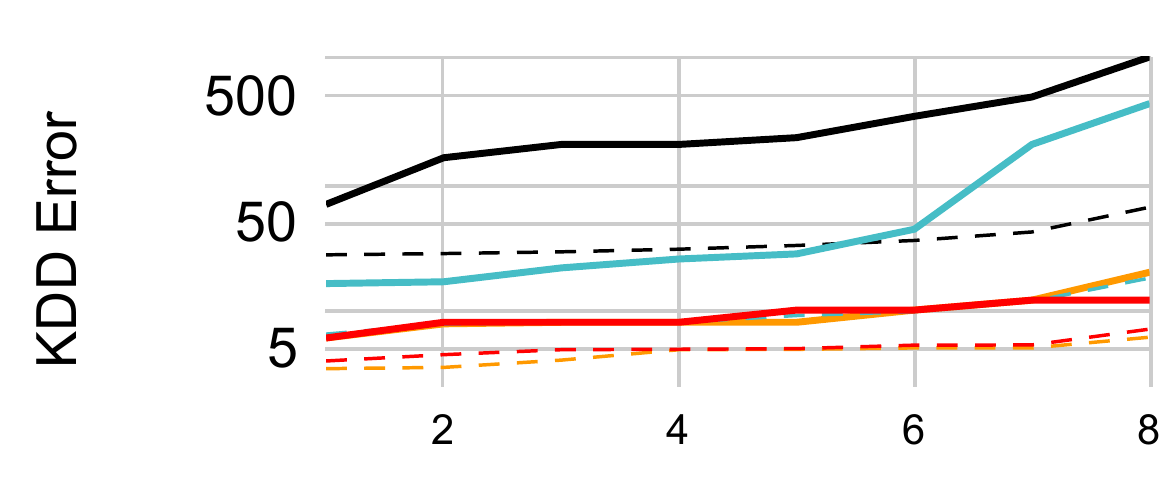}
\includegraphics[height=2.48cm]{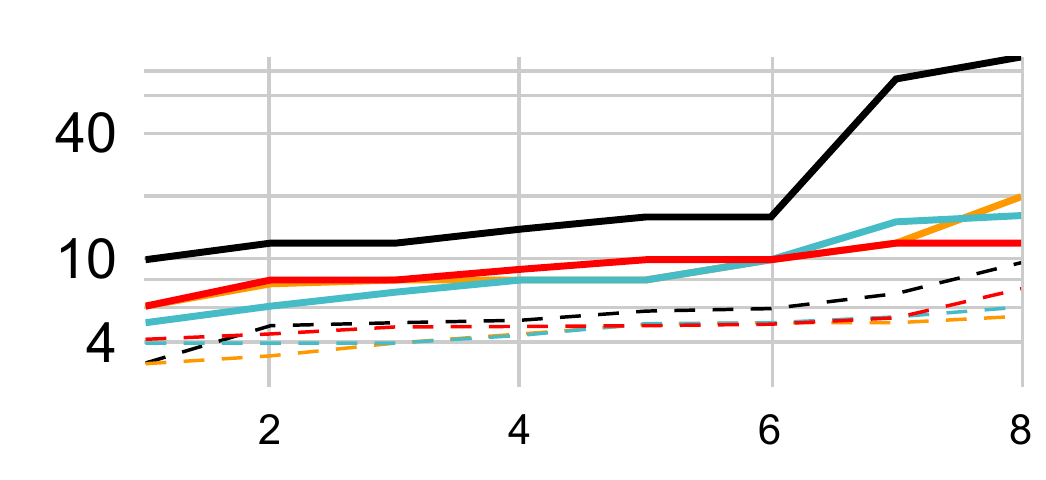}
\includegraphics[height=2.48cm]{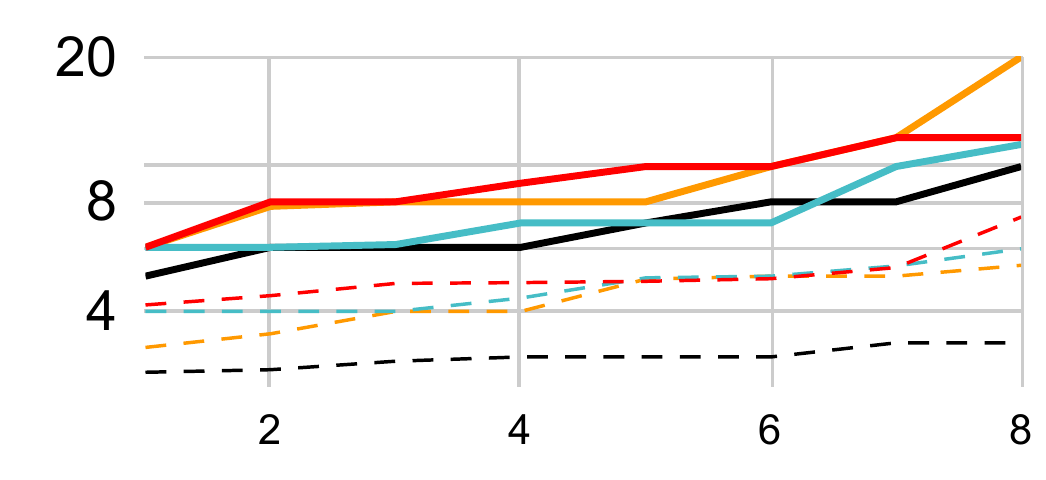}
\includegraphics[height=2.48cm]{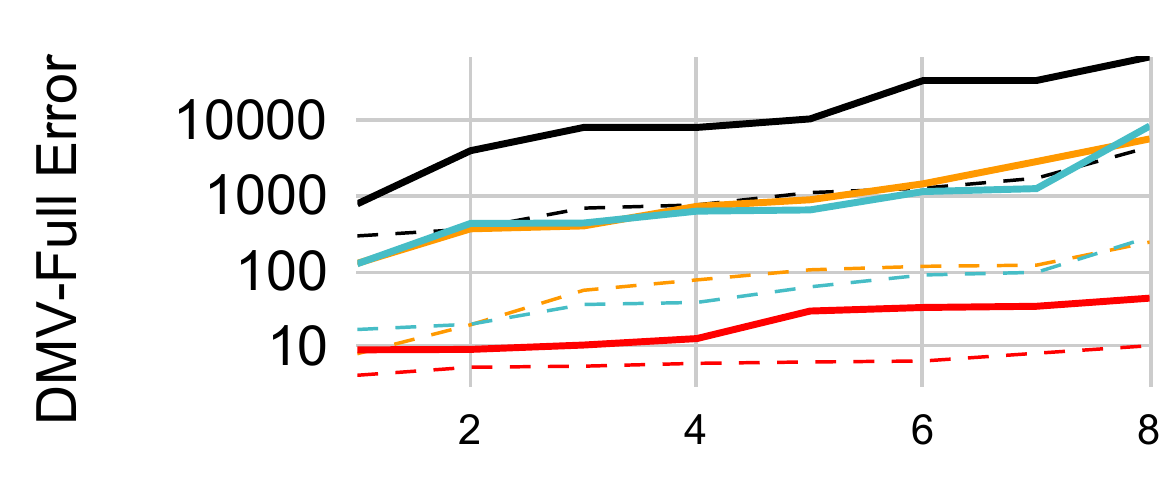}
\includegraphics[height=2.48cm]{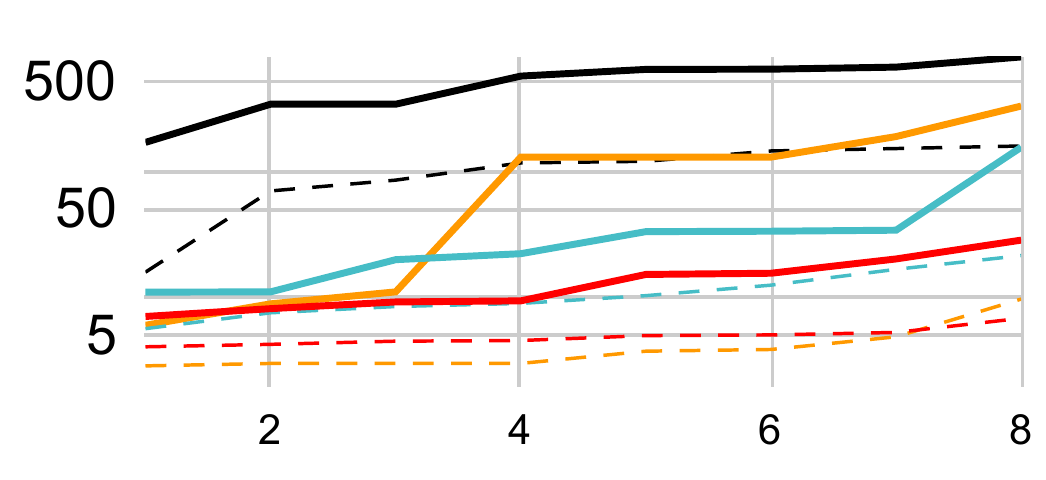}
\includegraphics[height=2.48cm]{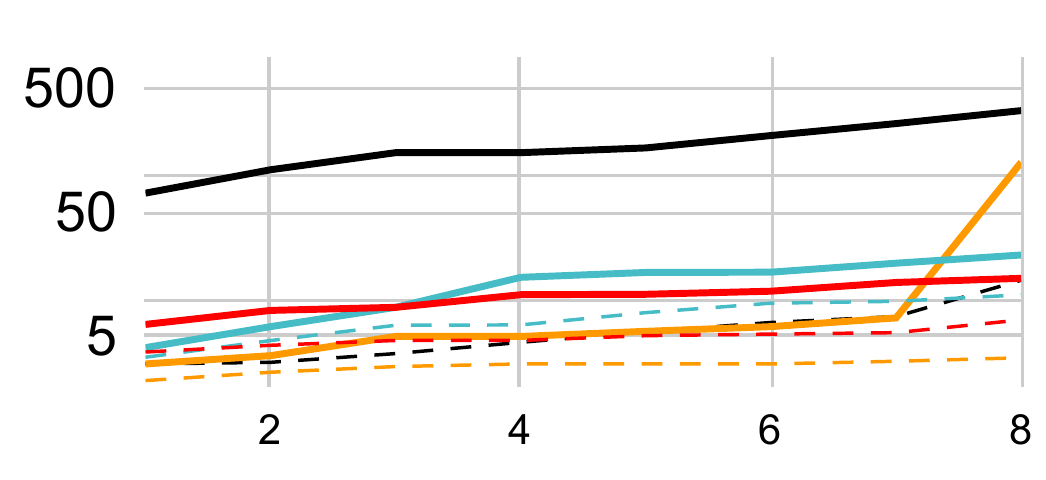}
\includegraphics[height=.4cm]{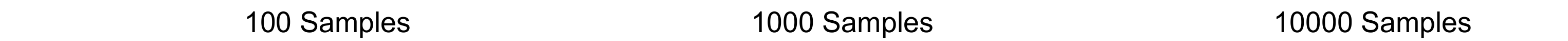}
\includegraphics[height=.48cm]{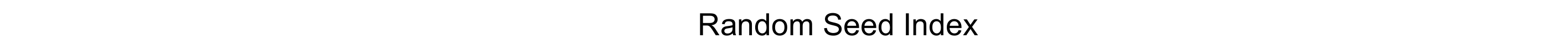}

\vspace{-.2cm}
\caption{Variable skipping and skipping combined with multi-order training vs. baselines across different datasets, variable orderings, and sampling budgets. Error is plotted on the y-axis in \textit{log scale} (lower is better). Each column reflects a $10\times$ increase in sampling budget as we move to the right. Results for 8 different variable orders within each plot are sorted by increasing error. Variable skipping provides $10{-}100\times$ max error reduction at low budgets, and still improves accuracy at high sampling budgets for large datasets such as \dmv. This data is also shown in tabular form in Table \ref{table:full}, which additionally reports median errors. \label{fig:main}}
\end{figure*}

\subsection{Evaluation Metric}
 We issue a large set of randomly generated range queries, and measure how accurately each estimator answers them.
 We report the multiplicative error, or \textit{Q-error}, defined as the factor by which an estimate differs from the actual density (obtained by actually executing each query on the dataset):
\begin{equation*}
\text{Error} := \max(\text{estimate}, \text{actual}) /  \min(\text{estimate}, \text{actual})
\end{equation*}
Hence, a perfect estimate for a query has an error of 1.0.
Moreover, we report the median, 99\%-tile, and maximum Q-error across all queries. We note that the median error is typically within a fraction of 1.0 for all estimators. The reason is that most randomly generated queries are ``easy'' (i.e., hit few cross-dimension predicate correlations), and only a few are ``hard''. Because of this, even a naive estimator can achieve good performance in many cases. Hence, our focus is on high quantile errors for evaluation.

\subsection{Experiment Setup}
\label{subsec:qgen}
For queries against tabular data, we used the experiment framework from \cite{naru}, randomly drawing between 5 and 12 conjunctive variable constraints per query\footnote{An example query for \dmv~may be: \texttt{{record\_type == 1 AND city == 17 AND zip > 10000 AND model\_year < 1990 AND max\_weight > 5000}}.}. It is important to not have too many or too few constraints, which would skew the distribution of true density estimates towards 0.0 (too many constraints lead to little density) or 1.0 (too few constraints lead to high density) respectively.

For text queries, we issued pattern glob queries of the form \texttt{value CONTAINS <str>}, where \texttt{<str>} is a character sequence between 3 and 5 characters in length drawn randomly from the full text corpus. This also provides a challenging spread of density from very common (e.g., \texttt{CONTAINS ".com"}), to quite rare (e.g., \texttt{CONTAINS "XVQ/i"}).


We compare between the following approaches, all of which use progressive sampling (Section~\ref{sec:prog-sample}) as the approximate inference procedure:
\begin{itemize}
\item \textbf{Baseline}: An autoregressive model queried using vanilla progressive sampling~\cite{naru}.
\item \textbf{Skipping}: An autoregressive model trained with random input masking and queried with the variable skipping optimization enabled (Section~\ref{sec:variable-skipping}).
\item \textbf{MultiOrder}: An autoregressive model trained under multiple variable orders to enable querying an ensemble of 10 orders at inference time~\cite{uria13_deep_tract_densit_estim,made}.
\item \textbf{MultiOrder + Skipping}: Combining the multi-order and variable skipping techniques.
\end{itemize}
The full list of training hyperparameters can be found in Table \ref{table:hyperparameters}.
Unless otherwise specified, we use a ResMADE \cite{resmade} with 3 residual blocks, two 256-unit hidden layers per block, and an 32-unit wide embedding for each input dimension. 
We choose hyperparameters known to optimize for progressive sampling performance \cite{naru}, but did not otherwise tune them for our experiments. In our ablations (Section \ref{subsec:ablation}) we found that the most sensitive hyperparameter to performance is the embedding size, which is closely related to model size. 

\begin{table*}[tp]
  \small
\centering
\caption{The model negative log-likelihoods at convergence in bits/datapoint (evaluated using non-masked data). We also report standard deviation across multiple random order seeds. 
 \label{table:config_bit_gap}}
\vspace{.1cm}
\begin{tabular}{l l l l l l} \toprule
{Dataset} & Baseline  & Random Input Masking & MultiOrder(5) & MultiOrder(10) & MultiOrder(15) \\ \midrule


\census & 52.04 $\pm$ .009 & {52.34} $\pm$ .02 & 52.69 $\pm$ .02 & 52.79 $\pm$ .03 & 52.81 $\pm$ .03\\
\dmv & 43.12 $\pm$ .04 & {43.65}  $\pm$ .06 & 44.15 $\pm$ .05 & 44.53 $\pm$ .06 & 44.65 $\pm$ .04\\
\kdd & 107.5 $\pm$ .3 & {116.58} $\pm$ .2 & 123 $\pm$ .9 & 127.6 $\pm$ .4 & 128.4 $\pm$ .5 \\

\bottomrule
\end{tabular}
\end{table*}
The autoregressive variable ordering can significantly affect estimator variance.
We thus evaluate each technique on 8 randomly chosen variable orderings and train a (fixed-order) model for each ordering\footnote{For ResMADE, this means we sample 8 sets of \{input ordering, intermediate connectivity masks\}.}.
For multi-order models, we train 8 distinct sets of 10 randomly chosen orders (we saw diminishing returns past 10 orders), unless specified otherwise.
To ensure fairness, when not using skipping, we use a model trained without masking.

For multi-order ResMADE, to condition on the current ordering statistics 
each masked linear layer is allocated an additional weight matrix that shares the existing mask and has an all-one vector as its input\footnote{This treatment has appeared in MADE~\cite{made}.}. Due to the additional weights, we size down the hidden units appropriately to ensure that the multi-order models have about the same parameter count as other models.


\subsection{Variable Skipping Performance}

We evaluate the impact of the variable skipping optimization on the \dmv, \census, and \kdd~datasets. For each dataset, we generated 1000 random range queries. 

In Figure \ref{fig:main} we show the results of variable skipping (orange and red lines) compared against baselines (black and turquoise lines). This data is also shown in tabular form in Table \ref{table:full}. We evaluate with sampling budgets of 100, 1000, and 10000 samples (left, center, and right columns respectively). Note that a sample refers to \textit{all} the forward passes required to sample relevant variables (e.g., for \census~a single sample takes 67 forward passes without skipping). We limit to 10k samples for cost reasons (at 10k samples, each query takes multiple seconds to evaluate even with a GPU). There are several key takeaways:

\textbf{High-quantile error differentiates estimators}: Across all estimators, the median error is very close to 1.0 (not shown since it is indistinguishable in log scale). However, systems applications necessarily seek to minimize the worst-case error, which does vary significantly across samplers.

\textbf{Skipping significantly improves sampling efficiency}:
Across all datasets, variable skipping provides between $10\times$ to $100\times$ max error reduction at low sampling budgets (i.e., 100 samples), compared to the baseline. It also provides up to $10\times$ improvement over the multi-order ensemble alone.

Concretely, at 100 samples the 99th-quantile error for \census~is reduced from $^\sim1000$ to $2.5$, \kdd~from $^\sim30$ to $3$, and \dmv~from $^\sim1000$ to $100$. Moving up to 1000 samples, we continue to see a significant improvement at the max error, with \census~improved from $^\sim500$ to $10$, \kdd~from $^\sim15$ to $8$, and \dmv~from $^\sim500$ to $100$.

Compared to multi-order, variable skipping provides $10\times$ better max error reduction for \census~and \kdd~at 100 samples. Interestingly, while multi-order and variable skipping provide comparable improvements for \dmv~at 100 samples, \textit{combining multi-order and skipping} provides a further $10\times$ improvement in both max and 99th quantile error. This suggests that variable skipping and multi-order training are orthogonal mechanisms, and can be combined for larger datasets such as \dmv~to reduce both error \textit{and} inference costs.



\textbf{Variable skipping can help even at high sampling budgets}: On the \dmv~dataset, variable skipping provides more than an order of magnitude reduction in max error (from $^\sim150$ to $5$), even at 10000 samples.
We hypothesize this is due to the large domain sizes of \dmv~(up to 32K distinct values), which in the worst case would require a much larger number of samples to achieve low estimation error.
As evidence for this, a large number of samples are required to achieve good errors even with skipping enabled. This is in contrast to the smaller \census~and \kdd~datasets where skipping achieves close to single-digit errors even with as low as 100 samples. This suggests that for even larger datasets (common in industrial settings), skipping may have an even greater impact.

We have also provided a summary of the results for Figure \ref{fig:main} in Figure \ref{fig:summary}. For the concrete target of $10\times$ error at the 99th quantile, skipping significantly reduces the compute requirements over both the baseline and multi-order. This is due to both accuracy improvements that combine with those provided by multi-order ensembles, and also the reduced compute requirements of skipping.


\subsection{Model Likelihoods vs. Training Scheme}

\begin{table*}[htp]
\caption{Variable skipping vs. vanilla progressive sampling on the text domain. Naive sampling refers to generating samples (from the learned AR model) without constraints and then filtering the generated samples to estimate the probability of matches. We include naive sampling as a baseline for this experiment since it is competitive with progressive sampling in the text domain. We measure the estimation error over 100 random pattern queries against the \dryad~dataset, and show the bootstrap standard deviation. \label{table:text}}
\vspace{.1cm}
\centering
\begin{tabular}{l l l l l} \toprule
Num Samples & Metric & Naive Sampling  & Progressive Sampling (Baseline) & Variable Skipping\\ \midrule

1000 & Max Error & 6412 $\pm$ 1280    & 4628 $\pm$ 1785
        & \textbf{115.2} $\pm$ 25.6\\
1000 & P99 Error & 4054 $\pm$ 1386    & 1741 $\pm$ 1824
        & \textbf{89.5} $\pm$ 30.6\\
1000 & Median Error & \textbf{1.23} $\pm$ .06   & 1.66 $\pm$ .15
        & 1.39 $\pm$ .08\\
\bottomrule
\end{tabular}
\end{table*}



Training with partially masked inputs makes the learning task more difficult: the number of examples increases by a factor of $2^N$, and effectively the same model is learning multiple autoregressive distributions.
Table~\ref{table:config_bit_gap} shows that in terms of negative log-likelihoods achieved, models trained with masking do have a slightly higher NLL than baseline, as expected.  However, the NLLs achieved are lower than those of multi-order models, and the gap only widens with an increased number of orders.

Even though NLLs of masked models are higher than those of baseline,
Figure~\ref{fig:main} shows the benefit: estimation error is significantly improved when variable skipping is enabled.
This highlights \emph{the non-perfect alignment of optimizing for point likelihoods vs. downstream range query performance}, opening up interesting future directions. 

\subsection{Model Size and Masking Ablations}
\label{subsec:ablation}


\begin{figure}
  \centering
\includegraphics[width=\columnwidth]{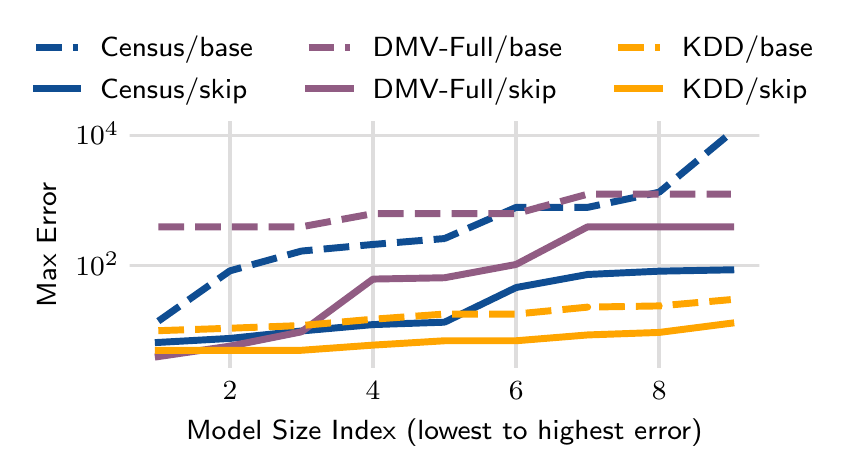}
\vspace{-.3in}
\caption{Model size vs. max estimation error on 1000 queries with 1000 samples each. The results for each dataset are sorted in increasing error. Errors are plotted on the y-axis in \textit{log scale} (lower is better). Errors increase as the model embedding sizes are reduced from 32 to 2, and hidden layer sizes from 256 to 16. Variable skipping (solid lines) retains an advantage across this two orders of magnitude change in model capacity.\label{fig:model-size}}
\end{figure}

In Figure \ref{fig:model-size} we study the relationship between model size and estimation accuracy. For this experiment, we vary the model embedding size among $\{2, 8, 32\}$ dimensions, and the hidden layer size among $\{16, 64, 256\}$ units. We see that variable skipping retains a robust advantage across nearly two orders of magnitude variation in model size.

\begin{figure}
  \centering
\includegraphics[width=\columnwidth]{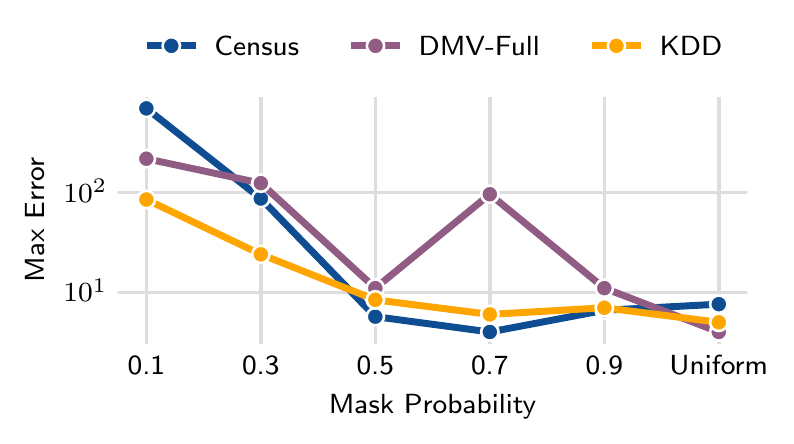}
\vspace{-.3in}
\caption{Varying the masking scheme. Here we measure the max estimator error with skipping enabled over 1000 queries with 1000 samples each, on the natural variable order. Errors are plotted on the y-axis in \textit{log scale} (lower is better). \label{fig:dropout-scheme}}
\end{figure}

In Figure \ref{fig:dropout-scheme} we compare a few schemes for selecting the mask distribution, from fixed masking probabilities of $\{0.1, 0.3, 0.5, 0.7, 0.9\}$, vs. the random uniform scheme used in the main experiment.
We see that drawing the mask probability uniformly at random obtains the lowest errors for \kdd~and \dmv, and close to optimal for \census~as well, showing it to be a robust choice.

\subsection{Application to Pattern Matching in Text Domain}
\label{subsec:texteval}

Finally, we show that variable skipping can be applied to the text domain for estimating the probability of pattern matches. Pattern matching (or more generally, regex matching), can be thought of as unrolling a dynamic predicate as variables are sampled (Section \ref{subsec:text}). Here we evaluate a simple character-level Transformer model on
the \dryad~dataset. 
We note that this is not a realistic application since scanning a dataset of this size is much faster than sampling from a model, however it demonstrates the applicability of variable skipping across domains. 
Table \ref{table:text} shows that prefix skipping enables much lower variance estimates than naive sampling and vanilla progressive sampling.

\section{Conclusion}
To summarize, we identify the range density estimation task and important applications. We propose \emph{variable skipping}, which greatly reduces sampling variance and inference latency. We validate the effectiveness of these techniques across a variety of datasets and model configurations.

\begin{table*}[ht]
\caption{The full table of quantiles across all random orders evaluated in Figure \ref{fig:main}. We show the mean and standard deviation of the quantiles across the random order seeds. \label{table:full}}
\vspace{.1cm}
\centering
\begin{tabular}{l l l r r r r} \toprule
Samples & Metric & Dataset & Progressive (Baseline) & Multi-Order & Skipping & Multi-Order + Skipping\\ \midrule

100 & P50 & Census &
   1.19 $\pm$ .01 &
   1.53 $\pm$ .32 &
   \textbf{1.09 $\pm$ .01} &
   1.21 $\pm$ .03\\

 &  & KDD &
   1.24 $\pm$ .02 &
   1.41 $\pm$ .16 &
   \textbf{1.14 $\pm$ .02} &
   1.25 $\pm$ .05\\
   
 &  & DMV-Full &
   1.22 $\pm$ .04 &
   1.70 $\pm$ .22 &
   \textbf{1.08 $\pm$ .01} &
   1.22 $\pm$ .04\\
   
 & P99 & Census &
   1150 $\pm$ 550 &
   38.6 $\pm$ 52 &
   \textbf{2.55 $\pm$ .29} &
   3.36 $\pm$ .32\\
   
 &  & KDD &
   36.4 $\pm$ 11 &
   9.82 $\pm$ 3.4 &
   \textbf{4.58 $\pm$ .83} &
   5.07 $\pm$ .85\\

 &  & DMV-Full &
   1130 $\pm$ 1300 &
   80 $\pm$ 81 &
   93 $\pm$ 69 &
   \textbf{6.5 $\pm$ 1.7}\\
   
 & Max & Census &
   24500 $\pm$ 18000 &
   3490 $\pm$ 6700 &
   15.1 $\pm$ 7.3 &
   \textbf{12.3 $\pm$ 5.3}\\
   
 &  & KDD &
   345 $\pm$ 280 &
   99.9 $\pm$ 140 &
   9.97 $\pm$ 4.1 &
   \textbf{9.25 $\pm$ 1.9}\\
   
 &  & DMV-Full &
   21200 $\pm$ 22000 &
   1640 $\pm$ 2600 &
   1560 $\pm$ 1700 &
   \textbf{22.7 $\pm$ 13}\\
  \hline
1000 & P50 & Census &
   \textbf{1.06 $\pm$ .01} &
   1.44 $\pm$ .31 &
   1.09 $\pm$ .01 &
   1.20 $\pm$ .02\\
   
 &  & KDD &
   \textbf{1.08 $\pm$ .01} &
   1.34 $\pm$ .17 &
   1.13 $\pm$ .02 &
   1.25 $\pm$ .05\\

 &  & DMV-Full &
   1.08 $\pm$ .01 &
   1.53 $\pm$ .24 &
   \textbf{1.05 $\pm$ .01} &
   1.18 $\pm$ .04\\
   
 & P99 & Census &
   2.58 $\pm$ .39 &
   3.84 $\pm$ 1.1 &
   \textbf{2.50 $\pm$ .28} &
   3.2 $\pm$ .28\\
   
 &  & KDD &
   5.78 $\pm$ 1.7 &
   4.69 $\pm$ .67 &
   \textbf{4.41 $\pm$ .74} &
   5.05 $\pm$ .89\\

 &  & DMV-Full &
   105 $\pm$ 44 &
   11.3 $\pm$ 4.8 &
   \textbf{4.24 $\pm$ 2.1} &
   4.92 $\pm$ .81\\

 & Max & Census &
   798 $\pm$ 700 &
   212 $\pm$ 330 &
   14.7 $\pm$ 7.0 &
   \textbf{12.8 $\pm$ 5.1}\\

 &  & KDD &
   30.7 $\pm$ 30 &
   9.42 $\pm$ 3.8 &
   9.95 $\pm$ 4.1 &
   \textbf{9.37 $\pm$ 1.9}\\

 &  & DMV-Full &
   508 $\pm$ 200 &
   39.4 $\pm$ 43 &
   114 $\pm$ 100 &
   \textbf{14 $\pm$ 6.7}\\
\hline

10000 & P50 & Census &
   \textbf{1.03 $\pm$ .01} &
   6.31 $\pm$ 1.6 &
   1.09 $\pm$ .01 &
   1.20 $\pm$ .24\\

 &  & KDD &
   \textbf{1.05 $\pm$ .01} &
   1.33 $\pm$ .18 &
   1.13 $\pm$ .02 &
   1.25 $\pm$ .05\\

 &  & DMV-Full &
   1.05 $\pm$ .01 &
   1.48 $\pm$ .24 &
   \textbf{1.05 $\pm$ .01} &
   1.18 $\pm$ .04\\
   
 & P99 & Census &
   \textbf{1.49 $\pm$ .08} &
   2.84 $\pm$ .70 &
   2.48 $\pm$ .29 &
   3.20 $\pm$ .29\\

 &  & KDD &
   \textbf{2.99 $\pm$ .19} &
   4.43 $\pm$ .93 &
   4.36 $\pm$ .75 &
   5.08 $\pm$ .85\\
   
 &  & DMV-Full &
   5.95 $\pm$ 3.4 &
   7.23 $\pm$ 2.4 &
   \textbf{2.89 $\pm$ .34} &
   4.96 $\pm$ .85\\
   
 & Max & Census &
   8.86 $\pm$ 14 &
   \textbf{6.31 $\pm$ 1.6} &
   14.7 $\pm$ 7.0 &
   12.6 $\pm$ 5.0\\

 &  & KDD &
   \textbf{7.0 $\pm$ 1.5} &
   7.57  $\pm$ 1.9 &
   9.97 $\pm$ 4.1 &
   9.37 $\pm$ 1.9\\

 &  & DMV-Full &
   181 $\pm$ 78 &
   13.5 $\pm$ 6.2 &
   20.2 $\pm$ 40 &
   \textbf{10.5 $\pm$ 2.6} \\

\bottomrule
\end{tabular}
\end{table*}


\bibliography{refs}
\bibliographystyle{icml2020}


\end{document}